\documentclass[10pt,twocolumn,letterpaper]{article}

\usepackage[pagenumbers]{cvpr} 
\usepackage{algorithmic}
\usepackage{algorithm}
\usepackage[table,xcdraw]{xcolor}
\usepackage{booktabs, multirow, makecell}
\usepackage{array}
\usepackage{nicematrix}
\usepackage{tikz}
\usepackage[accsupp]{axessibility} 
\definecolor{cvprblue}{rgb}{0.21,0.49,0.74}
\usepackage[pagebackref,breaklinks,colorlinks,allcolors=cvprblue]{hyperref}

\def\paperID{***} 
\def\confName{CVPR}
\def\confYear{2026}

\title{PAUL: Uncertainty-Guided Partition and Augmentation for Robust Cross-View Geo-Localization under Noisy Correspondence}

\author{
    Zheng Li$^1$ \quad Xueyi Zhang$^{2,3}$ \quad Yanming Guo$^1$ \quad Yuxiang Xie$^1$ \\
    Zhaoyun Ding$^1$ \quad Siqi Cai$^{2,4}$ \quad Haizhou Li$^{2,3}$ \quad Mingrui Lao$^{1}$\thanks{Corresponding author.} \\[6pt]
    $^1$National University of Defense Technology, Changsha, China \\
    $^2$Shenzhen Loop Area Institute, Shenzhen, China \\
    $^3$School of Artificial Intelligence, The Chinese University of Hong Kong, Shenzhen, China \\
    $^4$Harbin Institute of Technology, Shenzhen, China \\
     {\tt\small lizheng18@nudt.edu.cn \quad laomingrui@vip.sina.cn}
}

\begin{document}
\maketitle
\begin{abstract}
{Cross-view geo-localization (CVGL) is a critical task for UAV navigation, event detection, and aerial surveying, where existing approaches learn a shared embedding for drone and satellite images. However, these methods generally assume perfect alignment of image pairs in the training data, an assumption that rarely holds in practice. In real-world scenarios, factors such as urban canyon effects, electromagnetic interference, and adverse weather frequently induce GPS drift, resulting in systematic \textbf{alignment shifts} where only partial correspondences exist between image pairs. Despite its prevalence, the noisy correspondence issue in CVGL is still underexplored.
To address the gap, this work presents the first systematic investigation of the \textbf{Noisy Correspondence in Cross-View Geo-Localization (NC-CVGL)} problem, specifically focusing on the practical scenario where a significant portion of training pairs exhibit spatial misalignment due to GPS drift. To tackle this challenge, we propose \textbf{PAUL} (\textbf{P}artition and \textbf{A}ugmentation by \textbf{U}ncertainty \textbf{L}earning), a novel yet practical framework that achieves noise-robust learning through three coordinated mechanisms: \textbf{Co-partition} separates noisy from clean samples using loss patterns; \textbf{Co-augmentation} enhances high-confidence regions guided by uncertainty; and \textbf{Co-training} refines learning through an evidential regularizer.
Unlike prior methods that mainly filter noisy samples, PAUL leverages them effectively through this collaborative design. Comprehensive experiments validate the effectiveness of the proposed novel components in PAUL, which consistently outperforms other competitive noisy correspondence methods in various noise ratios.}
\end{abstract}

\section{Introduction}

\begin{figure}[!tb]
\centering
\vspace{-0.1in}
\includegraphics[width=0.99\linewidth]{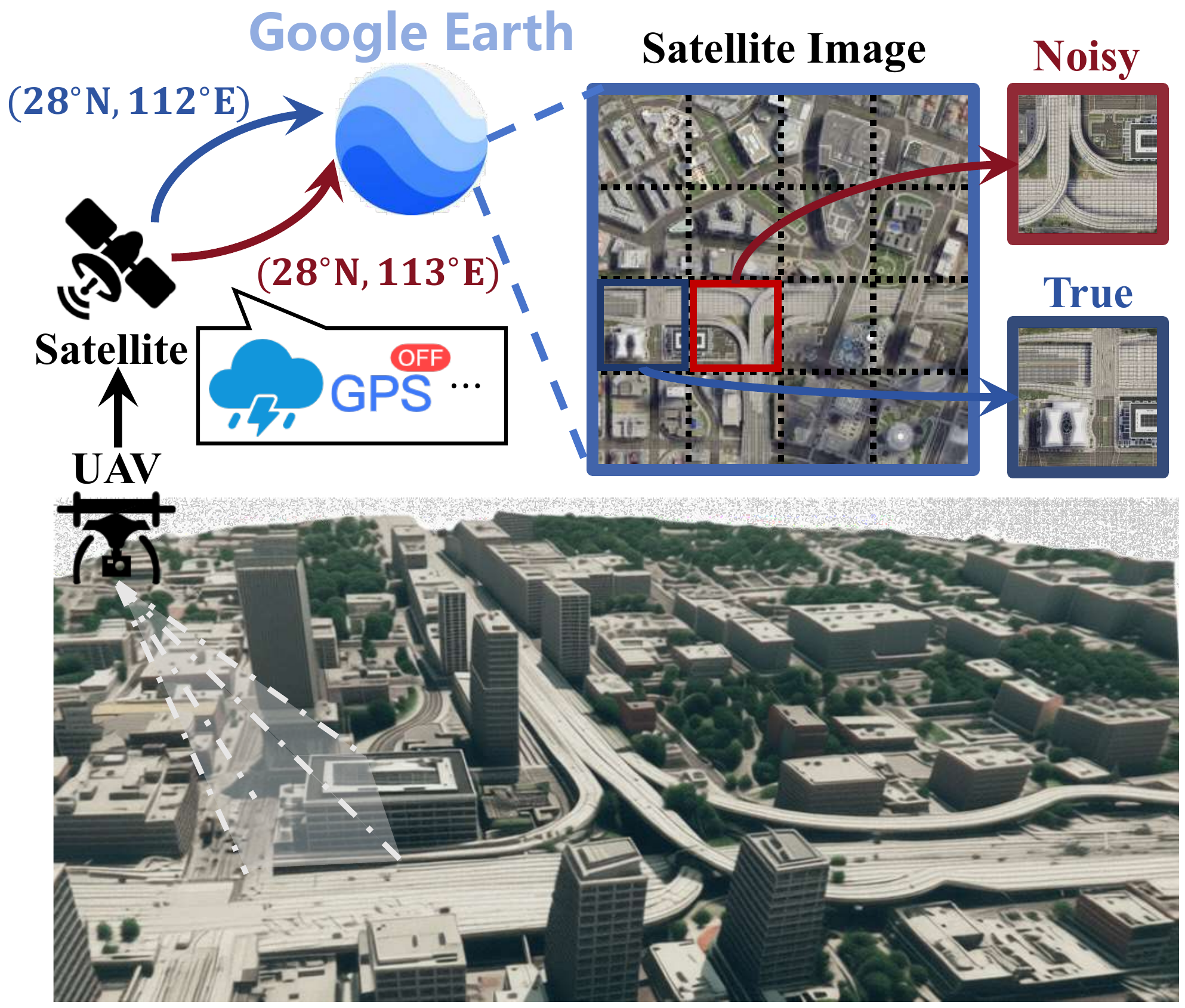}
\vspace{-0.05in}
\caption{\textbf{NC-CVGL}: Noisy correspondence resulting from inaccurate GPS information during data collection, leading to systematic misalignment between paired images.}
\vspace{-0.25in}
\label{fig:noisy-illustration}
\end{figure}

{UAV-Satellite Cross-View Geo-Localization (CVGL) aims to establish accurate correspondences between UAV-captured and satellite imagery for precise localization. This technology supports various UAV applications, including navigation, event detection, and aerial mapping~\cite{Multiple1, Multiple2, Multiple3}. To formulate the problem, existing advanced methods typically employ metric learning to match cross-view images by embedding them into shared feature spaces~\cite{sample4geo}. However, these approaches often rely on the assumption of perfect alignment and substantial spatial overlap in training image pairs, an idealization that rarely holds in practical scenarios.}



In real-world data collection for CVGL, multiple factors including urban canyons, electromagnetic interference, and adverse weather conditions frequently induce GPS drift. As illustrated in \cref{fig:noisy-illustration}, such coordinate inaccuracies result in systematic misalignments where satellite image tiles are offset from actual UAV capture locations. Consequently, many training pairs exhibit only partial spatial overlap despite semantic relevance, thereby degrading model performance.

To address this practical challenge, we formally define and investigate the \textbf{N}oisy \textbf{C}orrespondence in \textbf{CVGL} (\textbf{NC-CVGL}) problem, which simulates the impact of GPS drift in real deployment scenarios. This problem characterizes varying degrees of spatial misalignment in training pairs caused by coordinate inaccuracies. We quantify spatial correspondence quality using Intersection over Union (IoU), distinguishing well-aligned pairs from semi-positive noisy pairs with partial overlap. By tackling NC-CVGL, our work aims to mitigate the adverse effects of real-world GPS drift and enhance the robustness of cross-view geo-localization systems.

\begin{figure}[!h]
\centering
\includegraphics[width=0.99\linewidth]{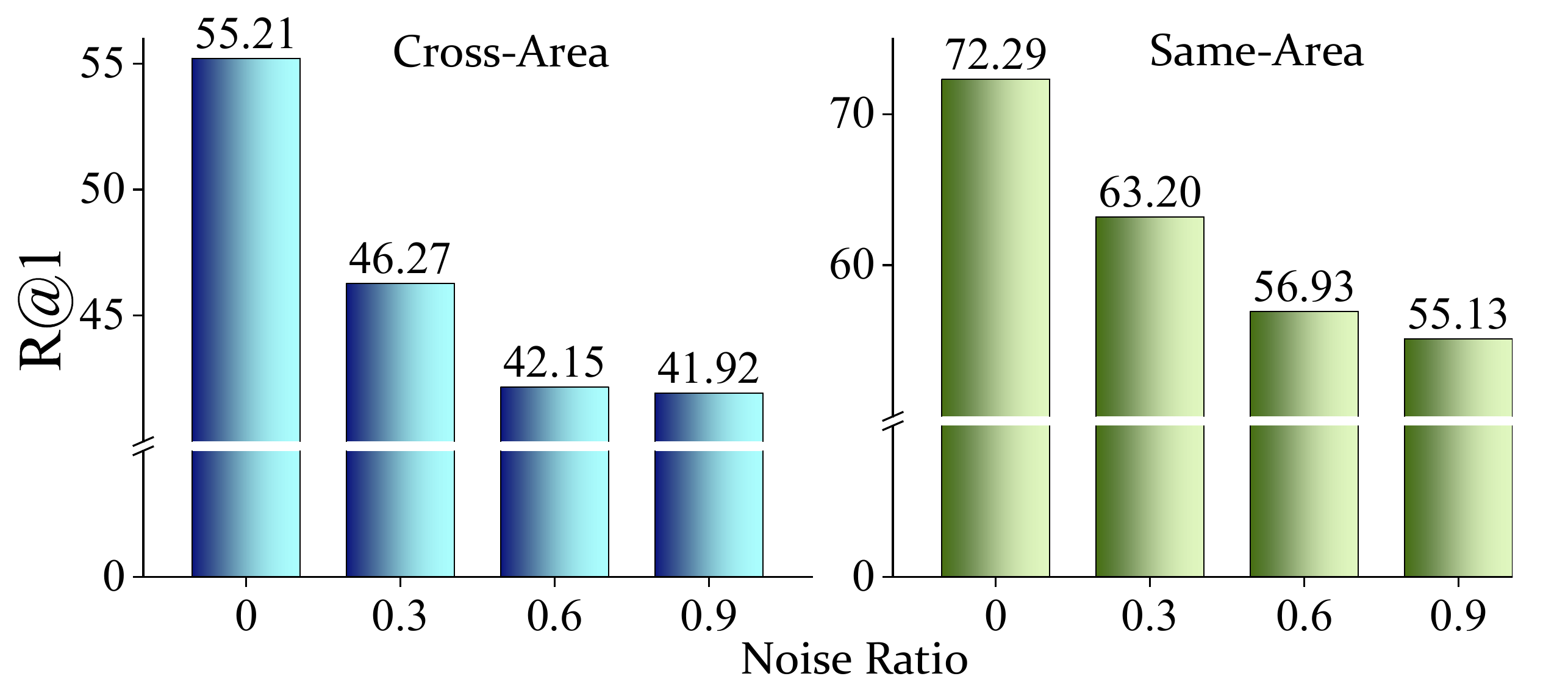}
\vspace{-0.1in}
\caption{State-of-the-art CVGL model Sample4Geo suffers noticeable performance degradation as the noise ratio increases.}
\vspace{-0.1in}
\label{fig:performance drop}
\end{figure}

NC-CVGL differs fundamentally from classical noisy correspondence problems, as it stems from coordinate inaccuracies causing systematic alignment bias rather than complete mismatches (\cref{fig:noisy-illustration}). While existing noisy correspondence methods either underutilize noisy data by discarding it entirely~\cite{MSCN,Bicro,ncr,dividemix} or repurpose it as negative samples or feature space anchors~\cite{gsc,esc}, such approaches prove suboptimal for NC-CVGL. The noisy pairs in NC-CVGL often retain substantial valuable information despite spatial misalignment, making their complete exclusion wasteful. This noise presents significant learning challenges: it forces nominally positive pairs to sustain high training loss and perturbs metric margins, substantially degrading model performance (\cref{fig:performance drop}). Although previous work addresses test-time viewpoint issues~\cite{sdpl}, the systematic correspondence noise induced by GPS drift during training remains under-explored, particularly in developing methods that can effectively leverage the informational content within noisy pairs.

{To address the challenges of NC-CVGL, we introduce \textbf{PAUL} (\textbf{P}artition and \textbf{A}ugmentation by \textbf{U}ncertainty \textbf{L}earning), a unified framework that transitions from aggressive noise filtering to latent signal exploitation. This approach enhances robustness through three systematically integrated components. First, \textbf{Co-partition} disentangles clean and noisy paris by analyzing the distinct learning dynamics of loss distributions. Subsequently, \textbf{Co-augmentation} rectifies the identified noisy pairs. Guided by evidential uncertainty cues, this module mines trustworthy local alignments to synthesize pseudo-clean supervision, thereby recovering valid signals from corrupted data. Finally, dual networks are optimized via \textbf{Co-training} with an evidential objective. This objective functions as a reliability-aware regularizer, attenuating the influence of unreliable samples by penalizing high-uncertainty predictions, while simultaneously capitalizing on the rectified information for superior representation learning.}


\textbf{Our main contributions are summarized as follows:}
\begin{itemize}
    \item {Task contribution:} To our best knowledge, this work serves as a pioneering study in systematically defining the NC-CVGL task and offering a theoretical analysis of alignment noise resulting from real-world GPS drift.
    \item {Methodological contribution:} We introduce the PAUL, grounded in uncertainty theory, which partitions and augments both ``clean” and ``noisy” data samples, extracts high-confidence information, and, by integrating evidential deep learning, achieves superior noisy data utilization and robustness compared to prior Noisy Correspondence methods;
    \item {Experimental contribution:} Extensive experiments on widely adopted datasets under various noise ratios demonstrate that PAUL achieves superior performance compared to other Noisy Correspondence methods.
\end{itemize}

\begin{figure*}[!tb]
    \centering    
    \includegraphics[width=0.9\linewidth]{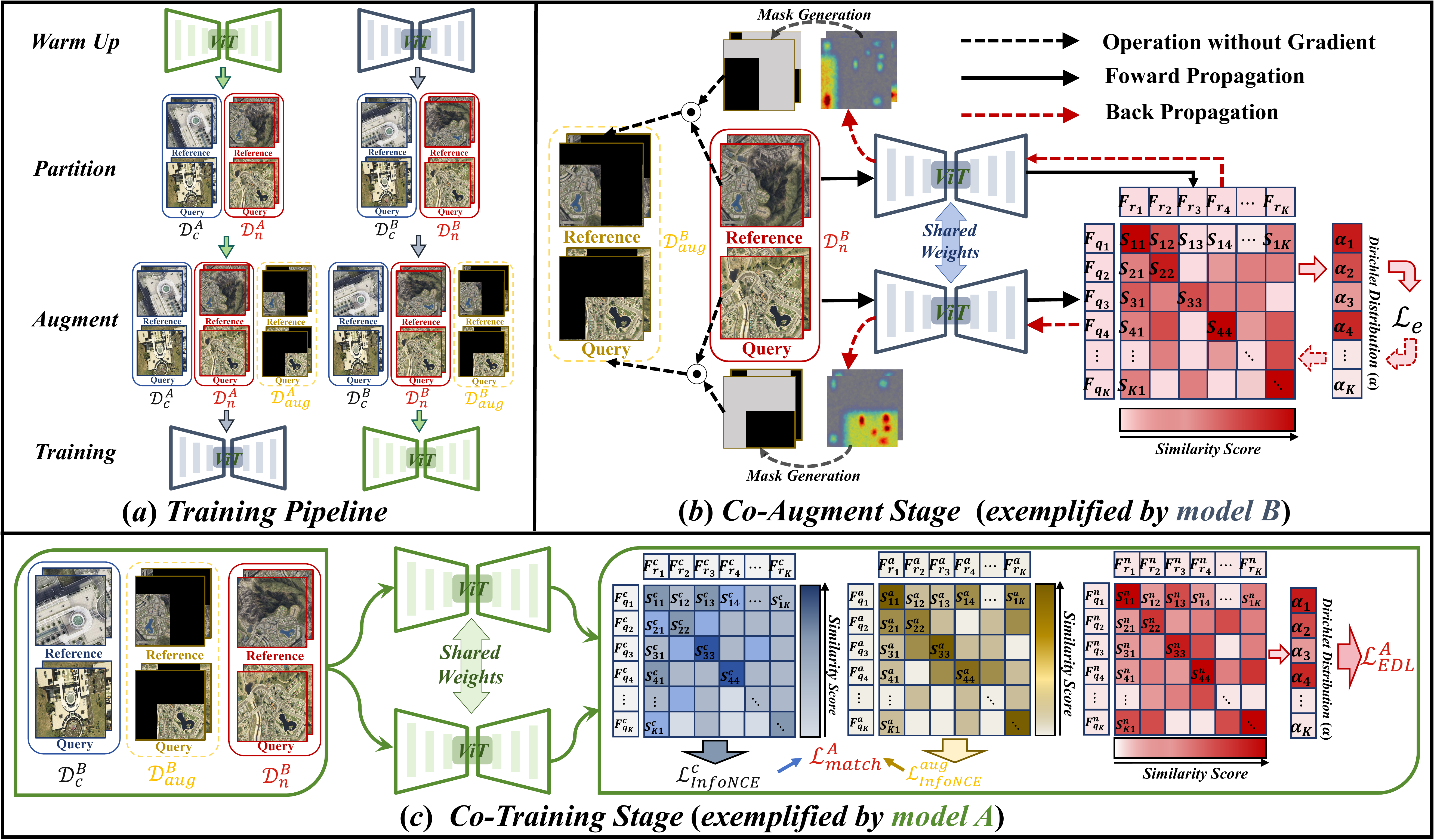} 
    \caption{
        Conceptual Illustration of PAUL: 
        \textbf{(a)} The complete framework, integrating both co-augmentation and co-training components.
        \textbf{(b)} Co-augmentation: noisy data are augmented based on EDL to enrich the training set.
        \textbf{(c)} Co-Training: two models are jointly optimized on the updated set, allowing mutual knowledge transfer and robustness to label noise.
    }
    \vspace{-0.2in}
    \label{fig:framework}
\end{figure*}

\section{Related Works}

\paragraph{Cross-View Geo-Localization}
Cross-view geo-localization (CVGL) aims to predict the geographic location of ground-level or UAV images by matching them to geo-tagged satellite views. Early handcrafted approaches~\cite{scale5,scale6,scale7} struggled with scale and viewpoint changes. Deep learning has since brought significant advances on benchmarks like CVUSA~\cite{scale35}, CVACT~\cite{scale36}, and Vigor~\cite{scale39}, Recent advancements include hard negative mining~\cite{sample4geo} and techniques such as unsupervised learning and cross-modal synthesis to reduce annotation costs and further narrow the domain gap between ground and satellite views~\cite{scale41,scale43,scale44}. UAV-specific datasets also begin to consider annotation noise and practical deployment scenarios~\cite{Multiple1,scale33,scale45,scale46}. Furthermore, structure-aware and part-based representations have been explored to improve robustness in challenging real-world conditions~\cite{scale50,sdpl}. \textit{However, most approaches still assume well-aligned training pairs, which is unrealistic for real UAV scenarios with frequent misalignment and noisy labels. To the best of our knowledge, our work is the first to formally describe this challenging task and to propose a tailored framework specifically designed to address this limitation. We emphasize that CVUSA, CVACT, and VIGOR are predominantly ground–satellite datasets, whereas our work explicitly targets the UAV–satellite setting: the query images are captured by UAV platforms and paired with nadir satellite tiles retrieved and cropped from a global/regional basemap using the UAV’s recorded GPS coordinates. This UAV–satellite formulation introduces distinct viewing geometry, scale and platform attitude effects, and a higher prevalence of GPS-induced correspondence noise, which our method is specifically designed to handle.}

\paragraph{Noisy Correspondence}
Learning from noisy correspondences is challenging in automatically labeled datasets~\cite{ncr}. Traditional strategies involve small-loss selection~\cite{MSCN,Bicro,ncr,dividemix}, while recent methods explore predictive inconsistency, uncertainty estimation~\cite{trip,cream,recon,UGNCL}, re-matching~\cite{l2rm}, pseudo-pairing~\cite{PC2}, robust objectives~\cite{rcl,decl,crcl}, or soft correspondence~\cite{esc,gsc}. Most focus on obvious mismatches or semantic noise, as in cross-modal retrieval. \textit{In CVGL, label noise is more subtle, typically spatial or partial misalignment, not semantic error. Despite its prevalence in real UAV data, most CVGL methods only improve robustness at inference~\cite{wang2024multiple,sdpl}, still relying on clean training data. To our knowledge, we are among the first to explicitly address noisy training correspondences in CVGL, enabling noise-resistant geo-matching.}

\paragraph{Data Uncertainty and Evidential Deep Learning}
Conventional neural networks lack explicit uncertainty modeling, which is problematic for noisy data~\cite{wen2023deep}. Uncertainty,  aleatoric or epistemic, can be estimated with deterministic schemes~\cite{NEURIPS2018_a981f2b7}, Bayesian inference~\cite{pmlr-v48-gal16, lao2023coca}, ensembling~\cite{NIPS2017_9ef2ed4b}, or test-time augmentation~\cite{pmlr-v124-lyzhov20a,gawlikowski2023survey}. Evidential Deep Learning~\cite{NEURIPS2018_a981f2b7} deterministically aggregates evidence and quantifies predictive vacuity. \textit{This is especially effective for identifying partially-aligned correspondences which is a key challenge in NC-CVGL.}

\section{Methodology}
\subsection{Problem Definition}
\begin{figure}[ht]
    \centering    
    \includegraphics[width=0.69\linewidth]{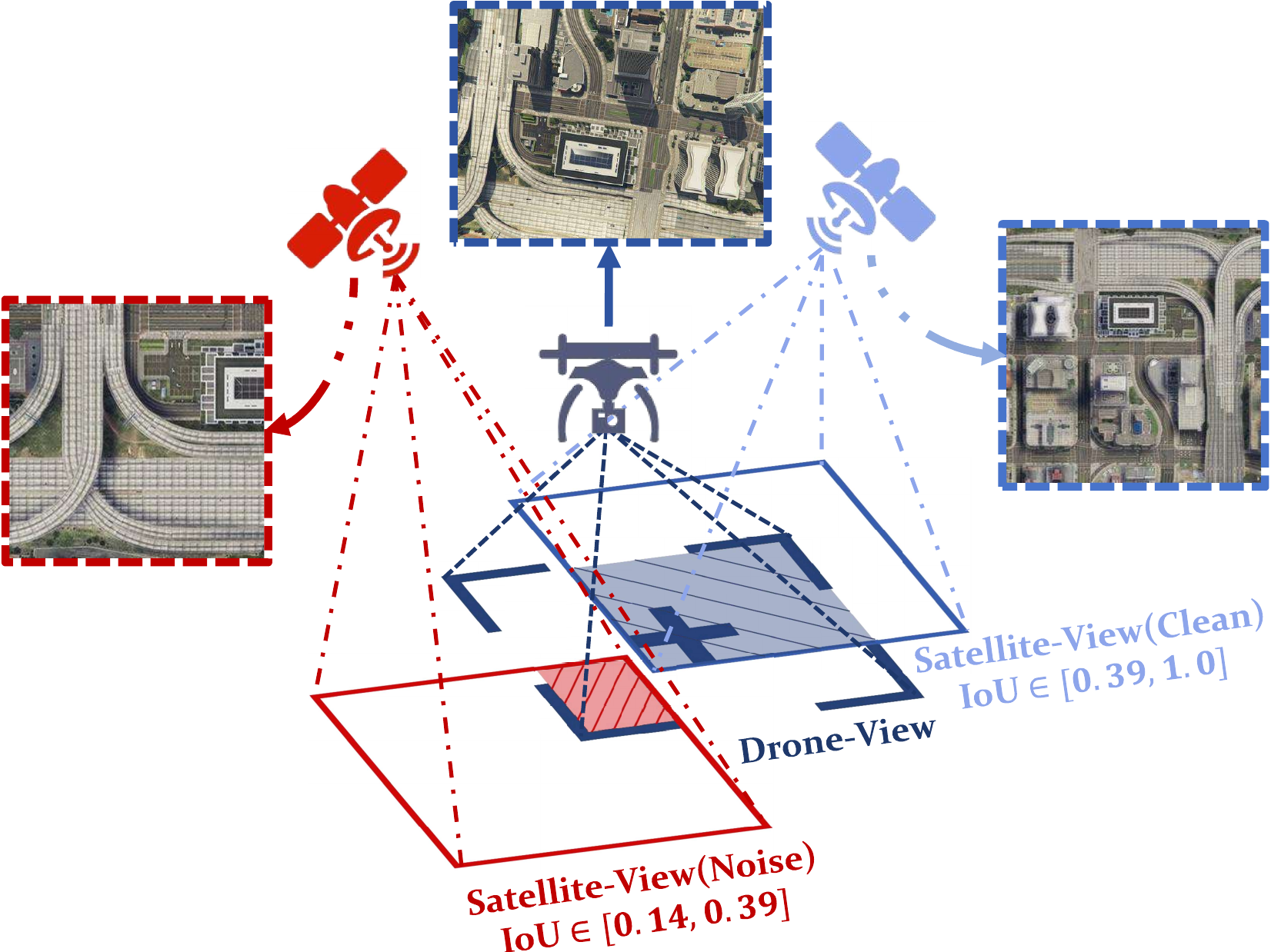} 
    \vspace{-0.1in}
    \caption{\textcolor{red}{Noisy samples} and \textcolor{blue}{clean samples} are divided based on their IoU with the drone-view images.}
    \vspace{-0.2in}
    \label{fig:IoU}
\end{figure}

In the Noisy Correspondence on Cross-View Geo-Localization (NC-CVGL), let $\mathcal{Q} = \{q_i\}_{i=1}^N$ be a set of UAV images and $\mathcal{R} = \{r_j\}_{j=1}^M$ a set of satellite images, each with geolocation annotations. Due to GPS drift or annotation errors, some pairs $(q_i, r_j)$ may be spatially misaligned. As shown in \cref{fig:IoU}, we measure their spatial overlap using Intersection over Union $\mathrm{IoU}(q_i, r_j)$, and use thresholds $\tau_m = 0.39$ and $\tau_s = 0.14$ to divide training pairs into:
\begin{align}
    \mathcal{P} &= \left\{ (q_i, r_j) \mid \mathrm{IoU}(q_i, r_j) > \tau_m \right\} \\
    \mathcal{N} &= \left\{ (q_i, r_j) \mid \tau_s < \mathrm{IoU}(q_i, r_j) \leq \tau_m \right\},
\end{align}
where $\mathcal{P}$ denotes well-aligned positive pairs, and $\mathcal{N}$ includes semi-positive pairs with spatial noise but partial overlap.
For each pair, $y_{ij} \in \{0, 1\}$ is the observed annotation indicating a matching pair, while $z_{ij} \in \{0, 1\}$ is a latent variable indicating whether the pair is a noisy semi-positive one (misaligned but overlapping). In practice, $y_{ij}$ is observed and used for training, but $z_{ij}$ is unobserved.
The goal is to learn an embedding function $f_\theta: \mathcal{Q}\cup\mathcal{R}\to\mathbb{R}^d$ such that matching pairs are mapped close in feature space. The training objective is:
\begin{equation}
    f_\theta^* = \arg\min_{f_\theta}\; 
    \mathbb{E}_{(q_i, r_j)\in \mathcal{P} \cup \mathcal{N}}
    \big[
        y_{ij}\cdot\ell(f_\theta(q_i), f_\theta(r_j), z_{ij})
    \big],
\end{equation}
where the loss $\ell$ depends on the (unobserved) noise indicator $z_{ij}$. Thus, the model must be robust to label noise arising from spatial misalignment.

\subsection{The Proposed Method} 
Motivated by the demand to fully utilize both clean and noisy correspondences while mitigating noise impact,  our proposed PAUL framework jointly leverages two independently and randomly initialized, identical neural networks and proceeds in three collaborative stages (\cref{fig:framework}(a)): (1) Co-partition: probabilistically partitioning clean and noisy pairs; (2) Co-augment: uncertainty-based spatial augmentation; (3) Co-training: dual-network supervision using exchanged datasets.

\subsubsection{Co-partition: Collaborative Division and Noise Controllable Sample Selection}

We base our model on the Sample4Geo framework. For a training pair $(q_i, r_j)$, the embedding features are extracted, and the InfoNCE loss is computed:
\begin{equation}
\label{eq:InfoNCE}
    \ell_{\mathrm{InfoNCE}} = -\log 
    \frac{\exp(S(q_i, r_j)/\tau)}
    {\sum_{k} \exp(S(q_i, r_k)/\tau)},
\end{equation}
where $S(\cdot)$ is feature similarity and $\tau$ is temperature.

Empirically, the distribution of $\ell_{\mathrm{InfoNCE}}$ over all training pairs naturally forms a mixture model, with low losses clustering for clean pairs (high IoU) and high losses for noisy pairs. This inspires us to adopt a soft and probabilistic partitioning by fitting a two-component Gaussian Mixture Model (GMM) over the loss values:
\begin{equation}
\label{eq:fitting GMM}
    p(\ell|\theta) = \beta\cdot \mathcal{N}(\ell; \mu_c, \sigma_c^2) + (1-\beta)\cdot \mathcal{N}(\ell; \mu_n, \sigma_n^2),
\end{equation}
where the first component models clean samples and the second noisy samples, and $\beta$ is the mixture weight. The EM algorithm is used to update these parameters. Each sample receives a posterior probability of being clean,
\begin{equation}
\label{eq:posterior of GMM}
    w_i = \frac{ \beta \cdot \mathcal{N} (\ell_i; \mu_c, \sigma_c^2)}
            { \beta \cdot \mathcal{N}(\ell_i; \mu_c, \sigma_c^2) + (1-\beta)\cdot \mathcal{N}(\ell_i; \mu_n, \sigma_n^2)}
\end{equation}
so that all pairs are simultaneously assigned soft membership to “clean” and “noisy” sets without arbitrary hard thresholds. This naturally allocates noisy (semi-positive) samples for further utilization rather than discarding them.

\subsubsection{Co-augment: Uncertainty-aware Region Masking via Evidential Deep Learning}

{To extract valuable local information from noisy pairs, we propose an uncertainty-aware masking mechanism via Evidential Deep Learning. This approach filters out high-uncertainty backgrounds to preserve only high-confidence regions, effectively distilling reliable signals from noise.} As illustrated in \cref{fig:framework}(b), to quantify uncertainty in the absence of ground-truth labels, we formulate the matching process within a batch of $K$ samples as a $K$-way classification problem. Specifically, for each sample $i$, the similarity logits $\boldsymbol{s}_i \in \mathbb{R}^K$ (derived from the $i$-th row of the similarity matrix) are transformed into an evidence vector $\mathbf{e}_i$:
\begin{equation}
\label{eq:definition of e}
    \mathbf{e}_i = \exp\left( \tanh\left(\frac{\boldsymbol{s}_i}{\tau}\right) \right),
\end{equation}
where $\tau$ is the temperature parameter. This evidence parameterizes a Dirichlet distribution with parameters $\boldsymbol{\alpha}_i = \mathbf{e}_i + 1$ and concentration strength $A_i = \sum_{k=1}^K \alpha_{ik}$. Consequently, the mean and variance of the pseudo-class probability predictions are derived as:
\begin{equation}
\label{eq:mean and variance}
    \mathbb{E}[p_{ik}] = \frac{\alpha_{ik}}{A_i}, \quad 
    \mathrm{Var}(p_{ik}) = \frac{\alpha_{ik}(A_i - \alpha_{ik})}{A_i^2(A_i + 1)}.
\end{equation}

To optimize this evidential belief, the EDL loss $\ell_\mathrm{EDL}$ is computed as:
\begin{align}
\label{eq:loss_edl}
    \ell_{\mathrm{EDL}} = \sum_{i=1}^{K} \Bigg\{ 
        &\sum_{k=1}^K \left[ 
            \left( y_{ik} - \frac{\alpha_{ik}}{A_i} \right)^2 
            + \frac{\alpha_{ik}(A_i - \alpha_{ik})}{A_i^2(A_i+1)} 
        \right] \notag \\
        &+ \lambda\, \mathrm{KL}\left( \mathrm{Dir}(\boldsymbol{\alpha}_i)\,\|\; \mathrm{Dir}(\mathbf{1}) \right)
    \Bigg\},
\end{align}
where the first term represents the Mean Squared Error (MSE) objective, and the second term acts as a Kullback-Leibler (KL) divergence regularizer enforcing a uniform prior. The total uncertainty for sample $i$ is quantified as $u_i = K/\left(\sum_k\alpha_{ik}\right)$.

Leveraging the interpretability of this formulation, we generate spatial masks to pinpoint critical regions via EDL-derived gradients. Specifically, we obtain the normalized activation map $H_x$ firstly:
\begin{equation}
\label{eq:Grad-CAM}
     H_x = \mathrm{Normalize}(\overline{G_x}), \; 
     G_x = \left| \frac{\partial \mathcal{L}_{\mathrm{EDL}}}{\partial x} \right|,
 \end{equation}
 where $\overline{G_x}$ denotes channel-wise mean pooling. To ensure the mask focuses on the most salient features, we apply a filtering and selection process:
 \begin{equation}
      \tilde{M}_x = \mathbb{I}[ H_x > \eta], \quad M_x = \mathcal{C}_{\mathrm{max}}(\tilde{M}_x),
 \end{equation}
 where $\eta$ is a threshold used to filter out low-response background noise. The operator $\mathcal{C}_{\mathrm{max}}(\cdot)$ extracts the largest connected component. This operation effectively removes ambiguous fragments and isolates the primary region of interest for robust model updates.

\subsubsection{Co-training: Cross-supervised Robust Learning}
{Despite data purification in prior stages, standard contrastive learning remains prone to confirmation bias from residual noise. To ensure robust convergence, we employ a Co-training strategy with an evidential regularizer, penalizing high-uncertainty predictions to suppress unreliable signals.} Specifically, following established noisy correspondence paradigms~\cite{ncr, Bicro}, we co-train two independent networks, A and B. At each iteration, the networks exchange supervision by sharing their respective sample selections (categorized as clean, augmented, or noisy) with their peer. For instance, the training set for model A is constructed using the partition derived from model B:

\begin{equation}
\label{eq:exchange dataset}
    \mathcal{D}^A = \mathcal{D}_c^B \cup \mathcal{D}_{aug}^B \cup \mathcal{D}_n^B.
\end{equation}

The optimization objective for network A is then decomposed into a standard matching term for reliable data and the proposed evidential term for the remaining noisy data:
\begin{align}
\label{eq:loss_tatol}
    \mathcal{L}_{\mathrm{total}}^A =\  
    &\underbrace{
        \sum_{(q_i, r_j) \in \mathcal{D}_c^B \cup \mathcal{D}_{aug}^B} 
        \ell_\mathrm{InfoNCE}(q_i, r_j)
    }_{\mathcal{L}_{match}} \notag \\
    &+ \lambda_\mathrm{EDL} \ 
    \underbrace{
        \sum_{(q_i, r_j) \in \mathcal{D}_n^B} 
        \ell_{\mathrm{EDL}}(q_i, r_j)
    }_{\mathcal{L}_\mathrm{EDL}},
\end{align}
where $\lambda_\mathrm{EDL}$ serves as a trade-off hyperparameter. 

\begin{algorithm}[!t]
\caption{Overview of PAUL}
\label{alg:pipeline}
\begin{algorithmic}[1]
\REQUIRE Query set $\mathcal{Q}$, Reference set $\mathcal{R}$, labels $y_{ij}$
\ENSURE Embedding networks $f_{\theta^A}, f_{\theta^B}$

\STATE \textbf{Initialize} parameters $\theta^A$, $\theta^B$ for $A,B$

\STATE \textbf{// Warmup}  
\FOR{warmup iteration}
    \FOR{$m \in \{A,B\}$}
        \STATE Update $\theta^m$ via InfoNCE (Eq.~\ref{eq:InfoNCE})
    \ENDFOR
\ENDFOR

\FOR{each epoch}
    \FOR{minibatch $\{(q_i, r_j)\}$}
        \FOR{$m \in \{A,B\}$}
            \STATE // \textcolor{cyan}{\textbf{Co-partition}}
            \STATE Compute InfoNCE loss $\ell_i^m$ for all pairs; fit GMM (Eq.~\ref{eq:fitting GMM}, \ref{eq:posterior of GMM}), partition $\mathcal{D}_c^m$, $\mathcal{D}_n^m$
            
            \STATE // \textcolor{cyan}{\textbf{Co-augment}}
            \FOR{$x_i^m \in \mathcal{D}_n^m$}
                \STATE Compute evidence $\mathbf{e}_i$ (Eq.~\ref{eq:definition of e}), Dirichlet $\boldsymbol{\alpha}_i$, and $\mathcal{L}_\mathrm{EDL}$ (Eq.~\ref{eq:loss_edl})
                \STATE Compute mask $M_x$ by saliency map (Eq.~\ref{eq:Grad-CAM}), mask low-confidence regions, add to $\mathcal{D}_{aug}^m$
            \ENDFOR
        \ENDFOR
        
        \STATE // \textcolor{cyan}{\textbf{Co-training}}
        \STATE $\mathcal{D}^A \leftarrow \mathcal{D}_c^B \cup \mathcal{D}_{aug}^B \cup \mathcal{D}_n^B$ 
        \STATE $\mathcal{D}^B \leftarrow \mathcal{D}_c^A \cup \mathcal{D}_{aug}^A \cup \mathcal{D}_n^A$~~(Eq.~\ref{eq:exchange dataset})
        \STATE Update $\theta^A$, $\theta^B$ using Eq.~\ref{eq:loss_tatol}
    \ENDFOR
\ENDFOR
\RETURN $f_{\theta^A}, f_{\theta^B}$
\end{algorithmic}
\end{algorithm}

In summary, \cref{alg:pipeline} outlines the complete workflow of the proposed PAUL framework. By adaptively partitioning training pairs via GMM-based loss statistics, rectifying noisy samples through uncertainty-aware region masking, and maximizing information recovery via dual-network co-training, our approach effectively tackles the challenges of NC-CVGL.

\begin{table*}[!ht]
\setlength{\tabcolsep}{2pt}
\aboverulesep=0ex
\belowrulesep=0ex
\caption{A comparison of performance on GTA-UAV with noise rates of 0.0\%, 30.0\%, and 60.0\%. The best and second-best results are marked in \textbf{bold} and \underline{underlined}, respectively.}
\label{tab:comparison}
\begin{NiceTabular}{c|c||ccccc||ccccc}
\toprule\toprule
\multirow{2}{*}{\textbf{Noise Ratio}} & \multirow{2}{*}{\textbf{Method}} 
  & \multicolumn{5}{c||}{\textbf{Cross-Area}}
  & \multicolumn{5}{c}{\textbf{Same-Area}} \\
\cmidrule{3-7} \cmidrule{8-12}
  &  & R@1 $\uparrow$ & R@5 $\uparrow$ & AP $\uparrow$ & SDM@3 $\uparrow$ & Dis@1 $\downarrow$ & R@1 $\uparrow$ & R@5 $\uparrow$ & AP $\uparrow$ & SDM@3 $\uparrow$ & Dis@1 $\downarrow$ \\
\midrule
\multirow{9}{*}{0.0\%}
 & InfoNCE & \cellcolor{gray!10}55.21\% & \cellcolor{gray!10}79.12\% & \cellcolor{gray!10}65.31\% & \cellcolor{gray!10}68.40\% & \cellcolor{gray!10}502.86m & \cellcolor{gray!10}\underline{72.29}\% & \cellcolor{gray!10}\underline{90.70}\% & \cellcolor{gray!10}80.46\% & \cellcolor{gray!10}\textbf{79.49}\% & \cellcolor{gray!10}738.49m \\
 & NCR     & \cellcolor{white}52.39\% & \cellcolor{white}74.73\% & \cellcolor{white}61.79\% & \cellcolor{white}62.34\% & \cellcolor{white}730.45m & \cellcolor{white}62.64\% & \cellcolor{white}82.44\% & \cellcolor{white}71.13\% & \cellcolor{white}70.15\% & \cellcolor{white}616.44m \\
 & BiCro   & \cellcolor{gray!10}43.73\% & \cellcolor{gray!10}68.71\% & \cellcolor{gray!10}54.40\% & \cellcolor{gray!10}57.43\% & \cellcolor{gray!10}894.75m & \cellcolor{gray!10}49.57\% & \cellcolor{gray!10}72.79\% & \cellcolor{gray!10}59.46\% & \cellcolor{gray!10}61.69\% & \cellcolor{gray!10}813.62m \\
 & CRCL    & \cellcolor{white}\underline{60.39}\% & \cellcolor{white}82.06\% & \cellcolor{white}69.66\% & \cellcolor{white}70.36\% & \cellcolor{white}443.22m & \cellcolor{white}49.01\% & \cellcolor{white}74.43\% & \cellcolor{white}59.74\% & \cellcolor{white}69.37\% & \cellcolor{white}738.60m \\
 & CREAM   & \cellcolor{gray!10}59.72\% & \cellcolor{gray!10}82.78\% & \cellcolor{gray!10}69.53\% & \cellcolor{gray!10}\underline{71.96}\% & \cellcolor{gray!10}\textbf{416.28m} & \cellcolor{gray!10}68.55\% & \cellcolor{gray!10}88.26\% & \cellcolor{gray!10}77.13\% & \cellcolor{gray!10}76.42\% & \cellcolor{gray!10}469.55m \\
 & ESC     & \cellcolor{white}39.75\% & \cellcolor{white}63.75\% & \cellcolor{white}49.87\% & \cellcolor{white}53.62\% & \cellcolor{white}996.10m & \cellcolor{white}48.72\% & \cellcolor{white}71.87\% & \cellcolor{white}58.47\% & \cellcolor{white}61.22\% & \cellcolor{white}839.42m \\
 & GSC     & \cellcolor{gray!10}59.36\% & \cellcolor{gray!10}\underline{82.77}\% & \cellcolor{gray!10}69.21\% & \cellcolor{gray!10}71.80\% & \cellcolor{gray!10}432.75m & \cellcolor{gray!10}72.11\% & \cellcolor{gray!10}90.67\% & \cellcolor{gray!10}\textbf{81.44}\% & \cellcolor{gray!10}79.40\% & \cellcolor{gray!10}\textbf{396.16m} \\
 & RCL     & \cellcolor{white}60.16\% & \cellcolor{white}82.70\% & \cellcolor{white}\underline{69.76}\% & \cellcolor{white}71.10\% & \cellcolor{white}473.10m & \cellcolor{white}71.08\% & \cellcolor{white}89.17\% & \cellcolor{white}79.01\% & \cellcolor{white}77.52\% & \cellcolor{white}469.62m \\
 & \textbf{PAUL}    & \cellcolor{blue!20}\textbf{61.21}\% & \cellcolor{blue!20}\textbf{83.59}\% & \cellcolor{blue!20}\textbf{70.66}\% & \cellcolor{blue!20}\textbf{72.06}\% & \cellcolor{blue!20}\underline{421.12m} & \cellcolor{blue!20}\textbf{73.55}\% & \cellcolor{blue!20}\textbf{91.17}\% & \cellcolor{blue!20}\underline{81.28}\% & \cellcolor{blue!20}\underline{79.41}\% & \cellcolor{blue!20}\underline{402.16m} \\
\midrule
\multirow{9}{*}{30.0\%}
 & InfoNCE & \cellcolor{white}46.27\% & \cellcolor{white}71.85\% & \cellcolor{white}57.14\% & \cellcolor{white}69.37\% & \cellcolor{white}494.98m & \cellcolor{white}63.20\% & \cellcolor{white}88.64\% & \cellcolor{white}74.28\% & \cellcolor{white}\textbf{82.66}\% & \cellcolor{white}639.92m \\
 & NCR     & \cellcolor{gray!10}44.40\% & \cellcolor{gray!10}66.79\% & \cellcolor{gray!10}53.79\% & \cellcolor{gray!10}59.05\% & \cellcolor{gray!10}890.46m & \cellcolor{gray!10}52.13\% & \cellcolor{gray!10}75.17\% & \cellcolor{gray!10}61.90\% & \cellcolor{gray!10}69.40\% & \cellcolor{gray!10}714.82m \\
 & BiCro   & \cellcolor{white}41.11\% & \cellcolor{white}67.20\% & \cellcolor{white}52.23\% & \cellcolor{white}58.61\% & \cellcolor{white}827.51m & \cellcolor{white}47.87\% & \cellcolor{white}73.73\% & \cellcolor{white}58.80\% & \cellcolor{white}68.16\% & \cellcolor{white}718.93m \\
 & CRCL    & \cellcolor{gray!10}53.79\% & \cellcolor{gray!10}79.13\% & \cellcolor{gray!10}64.58\% & \cellcolor{gray!10}72.46\% & \cellcolor{gray!10}449.21m & \cellcolor{gray!10}63.87\% & \cellcolor{gray!10}87.53\% & \cellcolor{gray!10}74.29\% & \cellcolor{gray!10}81.27\% & \cellcolor{gray!10}424.67m \\
 & CREAM   & \cellcolor{white}54.02\% & \cellcolor{white}\underline{80.48}\% & \cellcolor{white}\underline{65.20}\% & \cellcolor{white}\textbf{74.65}\% & \cellcolor{white}\underline{358.19m} & \cellcolor{white}64.22\% & \cellcolor{white}88.40\% & \cellcolor{white}{74.88}\% & \cellcolor{white}\underline{82.05}\% & \cellcolor{white}\underline{384.86m} \\
 & ESC     & \cellcolor{gray!10}40.84\% & \cellcolor{gray!10}67.44\% & \cellcolor{gray!10}52.18\% & \cellcolor{gray!10}59.41\% & \cellcolor{gray!10}803.40m & \cellcolor{gray!10}44.95\% & \cellcolor{gray!10}72.43\% & \cellcolor{gray!10}56.59\% & \cellcolor{gray!10}67.06\% & \cellcolor{gray!10}725.64m \\
 & GSC     & \cellcolor{white}\underline{54.41}\% & \cellcolor{white}80.20\% & \cellcolor{white}65.15\% & \cellcolor{white}73.49\% & \cellcolor{white}419.27m & \cellcolor{white}\underline{65.11}\% & \cellcolor{white}\textbf{88.53}\% & \cellcolor{white}\underline{75.32}\% & \cellcolor{white}82.02\% & \cellcolor{white}{396.23m} \\
 & RCL     & \cellcolor{gray!10}53.37\% & \cellcolor{gray!10}79.52\% & \cellcolor{gray!10}64.24\% & \cellcolor{gray!10}71.65\% & \cellcolor{gray!10}461.49m & \cellcolor{gray!10}64.67\% & \cellcolor{gray!10}86.79\% & \cellcolor{gray!10}74.29\% & \cellcolor{gray!10}79.88\% & \cellcolor{gray!10}459.39m \\
 & \textbf{PAUL}    & \cellcolor{blue!20}\textbf{58.70}\% & \cellcolor{blue!20}\textbf{82.13}\% & \cellcolor{blue!20}\textbf{68.74}\% & \cellcolor{blue!20}\underline{74.25}\% & \cellcolor{blue!20}\textbf{384.57m} & \cellcolor{blue!20}\textbf{68.52}\% & \cellcolor{blue!20}\underline{88.41}\% & \cellcolor{blue!20}\textbf{77.16}\% & \cellcolor{blue!20}{79.38}\% & \cellcolor{blue!20}\textbf{369.88m} \\
\midrule
\multirow{9}{*}{60.0\%}
 & InfoNCE & \cellcolor{gray!10}42.15\% & \cellcolor{gray!10}68.69\% & \cellcolor{gray!10}53.35\% & \cellcolor{gray!10}68.93\% & \cellcolor{gray!10}531.21m & \cellcolor{gray!10}56.93\% & \cellcolor{gray!10}84.73\% & \cellcolor{gray!10}69.15\% & \cellcolor{gray!10}82.61\% & \cellcolor{gray!10}375.96m \\
 & NCR     & \cellcolor{white}37.74\% & \cellcolor{white}59.74\% & \cellcolor{white}47.09\% & \cellcolor{white}54.37\% & \cellcolor{white}1070.31m & \cellcolor{white}47.48\% & \cellcolor{white}73.46\% & \cellcolor{white}58.59\% & \cellcolor{white}68.91\% & \cellcolor{white}732.37m \\
 & BiCro   & \cellcolor{gray!10}41.11\% & \cellcolor{gray!10}67.20\% & \cellcolor{gray!10}52.23\% & \cellcolor{gray!10}58.61\% & \cellcolor{gray!10}827.51m & \cellcolor{gray!10}47.31\% & \cellcolor{gray!10}74.99\% & \cellcolor{gray!10}58.95\% & \cellcolor{gray!10}70.16\% & \cellcolor{gray!10}634.73m \\
 & CRCL    & \cellcolor{white}48.20\% & \cellcolor{white}75.57\% & \cellcolor{white}59.92\% & \cellcolor{white}71.95\% & \cellcolor{white}502.51m & \cellcolor{white}59.72\% & \cellcolor{white}85.79\% & \cellcolor{white}71.00\% & \cellcolor{white}81.84\% & \cellcolor{white}{433.82m} \\
 & CREAM   & \cellcolor{gray!10}\underline{52.38}\% & \cellcolor{gray!10}\textbf{79.18}\% & \cellcolor{gray!10}\underline{63.52}\% & \cellcolor{gray!10}{73.32}\% & \cellcolor{gray!10}\textbf{395.64m} & \cellcolor{gray!10}58.61\% & \cellcolor{gray!10}{85.97}\% & \cellcolor{gray!10}{70.57}\% & \cellcolor{gray!10}\underline{82.77}\% & \cellcolor{gray!10}\textbf{353.71m} \\
 & ESC     & \cellcolor{white}44.42\% & \cellcolor{white}71.59\% & \cellcolor{white}55.97\% & \cellcolor{white}63.77\% & \cellcolor{white}697.04m & \cellcolor{white}48.57\% & \cellcolor{white}74.29\% & \cellcolor{white}59.44\% & \cellcolor{white}69.68\% & \cellcolor{white}683.12m \\
 & GSC     & \cellcolor{gray!10}51.20\% & \cellcolor{gray!10}77.36\% & \cellcolor{gray!10}62.13\% & \cellcolor{gray!10}\textbf{73.60}\% & \cellcolor{gray!10}{466.71m} & \cellcolor{gray!10}\underline{60.78}\% & \cellcolor{gray!10}\textbf{86.14}\% & \cellcolor{gray!10}\textbf{71.84}\% & \cellcolor{gray!10}{82.14}\% & \cellcolor{gray!10}430.12m \\
 & RCL     & \cellcolor{white}46.43\% & \cellcolor{white}72.99\% & \cellcolor{white}57.92\% & \cellcolor{white}70.50\% & \cellcolor{white}536.07m & \cellcolor{white}59.08\% & \cellcolor{white}84.44\% & \cellcolor{white}70.22\% & \cellcolor{white}79.78\% & \cellcolor{white}521.14m \\
 & \textbf{PAUL} & \cellcolor{blue!20}\textbf{52.61}\% & \cellcolor{blue!20}\underline{78.66}\% & \cellcolor{blue!20}\textbf{63.63}\% & \cellcolor{blue!20}\underline{73.39}\% & \cellcolor{blue!20}\underline{462.27m} & \cellcolor{blue!20}\textbf{61.31}\% & \cellcolor{blue!20}\underline{85.99}\% & \cellcolor{blue!20}\underline{71.76}\% & \cellcolor{blue!20}\textbf{82.84}\% & \cellcolor{blue!20}\underline{422.10m} \\
\bottomrule \bottomrule
\end{NiceTabular}
\vspace{-0.15in}
\end{table*}

\begin{table}[htbp]
\setlength{\abovecaptionskip}{5pt}
\setlength{\tabcolsep}{0.3pt}

\renewcommand\arraystretch{1}
\aboverulesep=0ex
\belowrulesep=0ex
\caption{A comparison of performance on UAV-VisLoc with noise rates of 0.0\%, 30.0\%. The best and second-best results are marked in \textbf{bold} and \underline{underlined}, respectively.}
\label{tab:visloc comparison}
\begin{NiceTabular}{c|c|ccccc}
\toprule\toprule
\textbf{Ratio} & \textbf{Method} & R@1 $\uparrow$ & R@5 $\uparrow$ & AP $\uparrow$ & SDM@3 $\uparrow$ & Dis@1 $\downarrow$ \\
\midrule
\multirow{8}{*}{0.0\%}
 & InfoNCE & 33.24\% & 52.07\% & 41.91\% & 48.93\% & 1451.71m \\
 & NCR     & \cellcolor{gray!10}24.57\% & \cellcolor{gray!10}42.06\% & \cellcolor{gray!10}33.42\% & \cellcolor{gray!10}42.91\% & \cellcolor{gray!10}1707.18m \\
 & BiCro   & 27.10\% & 46.19\% & 35.69\% & 44.91\% & 1638.99m \\
 & CRCL    & \cellcolor{gray!10}33.18\% & \cellcolor{gray!10}53.21\% & \cellcolor{gray!10}43.25\% & \cellcolor{gray!10}\textbf{52.99}\% & \cellcolor{gray!10}\underline{1359.46m} \\
 & CREAM   & 29.64\% & 50.07\% & 39.22\% & 49.49\% & 1432.41m \\
 & ESC     & \cellcolor{gray!10}27.64\% & \cellcolor{gray!10}47.53\% & \cellcolor{gray!10}37.07\% & \cellcolor{gray!10}47.83\% & \cellcolor{gray!10}1509.22m \\
 & GSC     & \underline{33.72}\% & \underline{53.55}\% & \underline{43.38}\% & 51.51\% & 1368.72m \\
 & \textbf{PAUL}    & \cellcolor{blue!20}\textbf{36.12}\% & \cellcolor{blue!20}\textbf{54.74}\% & \cellcolor{blue!20}\textbf{44.97}\% & \cellcolor{blue!20}\underline{52.05}\% & \cellcolor{blue!20}\textbf{1353.95m} \\
\bottomrule
\multirow{8}{*}{30.0\%}
 & InfoNCE & \underline{24.57}\% & 42.59\% & 33.53\% & 48.91\% & 1464.49m \\
 & NCR     & \cellcolor{gray!10}16.15\% & \cellcolor{gray!10}32.31\% & \cellcolor{gray!10}24.11\% & \cellcolor{gray!10}36.96\% & \cellcolor{gray!10}2020.24m \\
 & BiCro   & 18.96\% & 36.05\% & 26.68\% & 40.93\% & 1797.88m \\
 & CRCL    & \cellcolor{gray!10}23.77\% & \cellcolor{gray!10}\underline{44.19}\% & \cellcolor{gray!10}\underline{33.64}\% & \cellcolor{gray!10}\textbf{50.07}\% & \cellcolor{gray!10}\textbf{1325.91m} \\
 & CREAM   & 19.23\% & 35.65\% & 26.82\% & 43.21\% & 1648.69m \\
 & ESC     & \cellcolor{gray!10}18.83\% & \cellcolor{gray!10}37.65\% & \cellcolor{gray!10}27.39\% & \cellcolor{gray!10}38.36\% & \cellcolor{gray!10}1898.57m \\
 & GSC     & 21.09\% & 41.39\% & 30.98\% & 48.32\% & 1428.73m \\
 & \textbf{PAUL}    & \cellcolor{blue!20}\textbf{26.64}\% & \cellcolor{blue!20}\textbf{46.80}\% & \cellcolor{blue!20}\textbf{36.37}\% & \cellcolor{blue!20}\underline{49.44}\% & \cellcolor{blue!20}\underline{1382.79m} \\
\bottomrule\bottomrule
\end{NiceTabular}
\vspace{-0.2in}
\end{table}

\section{Experiments}

\subsection{Datasets}
We evaluate our method on two benchmark datasets: the large-scale, publicly available synthetic GTA-UAV dataset~\cite{scale46}, and a real-world dataset that we construct based on the UAV-VisLoc dataset~\cite{VisLoc}. The GTA-UAV dataset, consisting of 33,763 UAV images with precise GPS annotations, provides both positive and semi-positive samples, enabling comprehensive evaluation in both cross-area and same-area splits. \textit{Notably, to the best of our knowledge, GTA-UAV is currently the only publicly available dataset that contains both positive and semi-positive samples suitable for the NC-CVGL task.} The UAV-VisLoc dataset contains 6,742 pairs of nadir UAV images and corresponding high-resolution satellite maps across 11 regions in China. While the dataset provides only geographic coordinates without fine-grained correspondence, we reorganize the data following the GTA-UAV protocol~\cite{scale46}: images from Changjiang-20 and Taizhou-1 are used for training, and Changjiang-23 and Taizhou-6 are reserved for cross-area testing, ensuring no spatial overlap. This partitioning facilitates rigorous and fair real-world evaluation.
\subsection{Evaluation Protocols}
Consistent with established protocols~\cite{scale39, scale45, Multiple1}, for each UAV query image, we retrieve the top $K$ most similar geo-referenced satellite images. Evaluation metrics include Recall@$K$ (R@$K$), Average Precision (AP), SDM@$K$, and top-1 localization error (Dis@1).

\subsection{Implementation Details}
We adopt ViT-Base~\cite{sdpl43} as the encoder for both UAV and satellite images, with all inputs resized to $384\times384$. Training follows Sample4Geo~\cite{sample4geo}: Adam optimizer(initial LR $1\mathrm{e}{-4}$, cosine scheduling), 5 epochs, batch size 64, and 1 warmup epoch with basic training only. All experiments are conducted on a single NVIDIA RTX 3090 GPU.

\subsection{Comparisons with the State-of-the-Art}
We compare our method (PAUL) with recent baselines: InfoNCE~\cite{sample4geo}, NCR~\cite{ncr}, BiCro~\cite{Bicro}, RCL~\cite{rcl}, CRCL~\cite{crcl}, ESC~\cite{esc}, GSC~\cite{gsc}, and CREAM~\cite{cream}. On GTA-UAV, experiments are conducted under noise ratios of 0\%, 30\%, and 60\%; for UAV-VisLoc, noise ratios are 0\% and 30\%. Results are reported in \cref{tab:comparison} and \cref{tab:visloc comparison}.


Our experimental results demonstrate that PAUL consistently surpasses all baseline methods across various configurations, achieving the highest R@1 scores in all settings and exhibiting strong robustness and effectiveness in handling noisy correspondences. This superior performance originates from PAUL's dedicated mechanisms for effectively leveraging noisy samples: for misaligned pairs, we implement targeted data augmentation strategies to enhance their matching consistency and informational value, thereby amplifying beneficial signals concealed within noisy samples. Furthermore, the integration of evidential learning empowers the model to assess the reliability of each noisy pair, adaptively mitigating overfitting to unreliable samples while preserving the ability to extract discriminative features. These coordinated mechanisms ensure that PAUL not only avoids discarding noisy data but actively enhances their utility while minimizing adverse effects, ultimately yielding a more robust and noise-resistant representation space compared to conventional approaches.

\subsection{Ablation Study}
We perform ablation studies to validate each component and critical design choice in PAUL.

\subsubsection{Effectiveness of Different Components}
\cref{tab:ablation} shows that removing either the $\mathcal{L}_{match}$ or the $\mathcal{L}_{EDL}$ leads to a clear drop in retrieval performance. This demonstrates that both enhanced contrastive supervision and uncertainty modeling contribute independently and complementarily to learning robustness. Specifically, $\mathcal{L}_{match}$ ensures that reliable correspondences effectively guide representation learning, while $\mathcal{L}_\mathrm{EDL}$ enables the model to exploit noisy pairs by focusing on confident regions and mitigating the influence of ambiguous noise. Their integration forms a cooperative learning mechanism: clean supervision stabilizes feature space, and uncertainty-driven learning further utilizes difficult samples for stronger generalization under high noise. Thus, the ablation results reflect that both components are indispensable and together yield synergistic gains in robustness.

\begin{table}[!htbp]
\setlength{\tabcolsep}{1pt}

\centering
\aboverulesep=0ex
\belowrulesep=0ex
\caption{Ablation study of different components in PAUL on the GTA-UAV cross-area split with a noise rate of 30.0\%. The best results are highlighted in \textbf{bold}.}
\vspace{-0.1in}
\label{tab:ablation}
\rowcolors{2}{gray!10}{white} 
\begin{NiceTabular}{cc||ccccc}
\toprule \toprule
$\mathcal{L}_{match}$ & $\mathcal{L}_{EDL}$ & R@1 $\uparrow$ & R@5 $\uparrow$ & AP $\uparrow$ & SDM@3 $\uparrow$ & Dis@1 $\downarrow$  \\
\midrule
                  &                   & 46.27\%   & 71.85\%   & 57.14\%   & 69.37\%   & 494.98m \\
\cellcolor{gray!10}\checkmark        &         \cellcolor{gray!10}          & \cellcolor{gray!10}{55.72}\%   & \cellcolor{gray!10}{80.19}\%   & \cellcolor{gray!10}{66.04}\%   & \cellcolor{gray!10}{73.65}\%   & \cellcolor{gray!10}433.96m \\
                  & \checkmark        & 54.36\%   & 79.40\%   & 64.96\%   & 73.13\%   & {430.39m} \\
\rowcolor{blue!20} 
\checkmark        & \checkmark        & \textbf{58.70}\%   & \textbf{82.13}\%   & \textbf{68.74}\%   & \textbf{74.25}\%   & \textbf{384.57m} \\
\bottomrule \bottomrule
\end{NiceTabular}
\end{table}

\subsubsection{Analysis of Hyperparameter Sensitivity}
\begin{figure}[htbp]
    \centering
    \vspace{-0.05in}
    \includegraphics[width=0.99\linewidth]{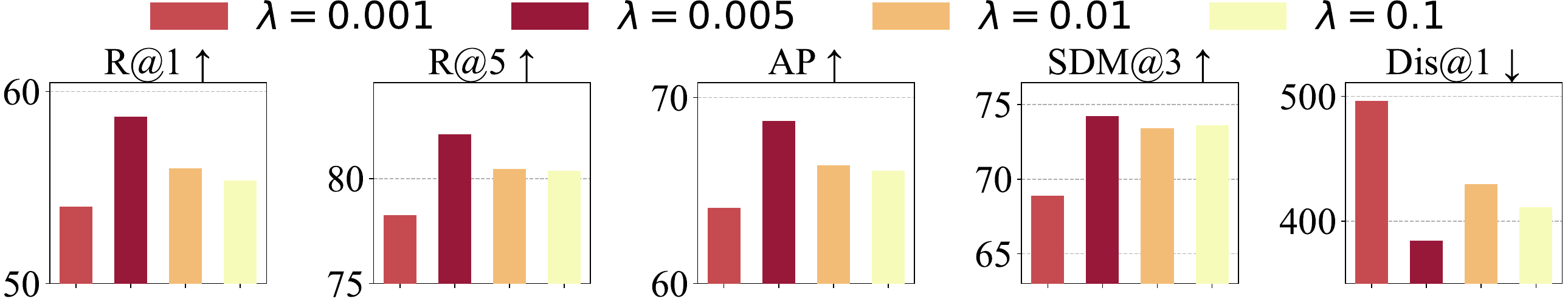} 
    \vspace{-0.05in}
    \caption{Performance of different values of $\lambda$ on GTA-UAV (cross-area, 30\% noise).}
    \label{fig:lambda}
\end{figure}

\begin{figure}[htbp]
    \centering
    \vspace{-0.05in}
    \includegraphics[width=0.99\linewidth]{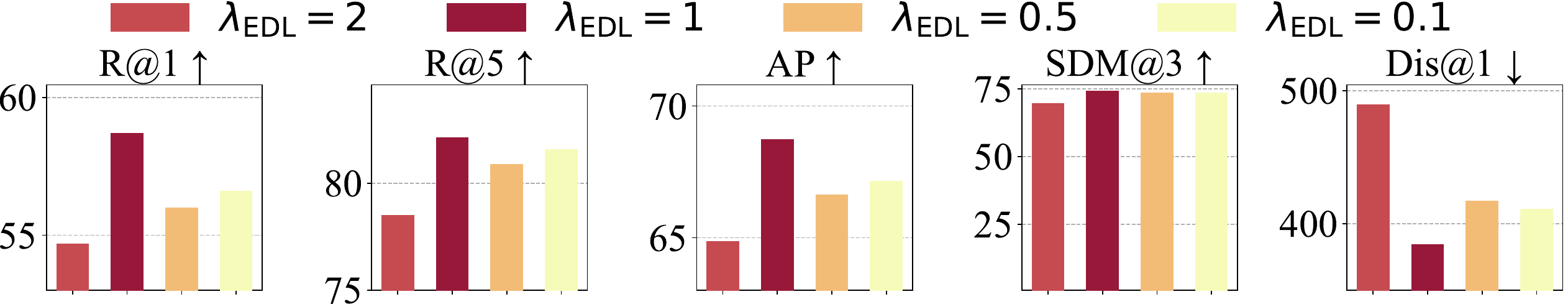} 
    \vspace{-0.05in}
    \caption{Performance of different values of $\lambda_{\text{EDL}}$ on GTA-UAV (cross-area, 30\% noise).}
    \label{fig:lambda-edl}
\end{figure}

We investigate the sensitivity of our method to key hyperparameters using the GTA-UAV dataset under the cross-area split with a noise ratio of 30\%. In particular, we analyze the impact of the trade-off hyperparameters $\lambda$($\ell_{\mathrm{EDL}}$'s balancing coefficient) and $\lambda_{\text{EDL}}$($\mathcal{L}_{\text{total}}$'s trade-off hyperparameter) on performance, as shown in \cref{fig:lambda} and \cref{fig:lambda-edl}.

\textbf{Effect of $\lambda$.} \cref{fig:lambda} presents the results for different values of $\lambda$. We observe that increasing $\lambda$ from 0.001 to 0.005 leads to clear gains across the board. However, further increases either plateau or slightly diminish the results, suggesting that finely tuning this parameter is important for balancing the main loss during training.

\textbf{Effect of $\lambda_{\text{EDL}}$.} Similarly, in \cref{fig:lambda-edl}, we analyze the effect of varying $\lambda_{\text{EDL}}$. The performance improves as $\lambda_{\text{EDL}}$ increases to 1, but further increasing this value leads to decreased results. This demonstrates that appropriate regularization from the uncertainty loss is beneficial, but excessive regularization can be detrimental.

\subsubsection{Analysis of Data Partition Strategy}
\begin{figure}[htbp]
    \centering
    \vspace{-0.05in}
    \includegraphics[width=0.99\linewidth]{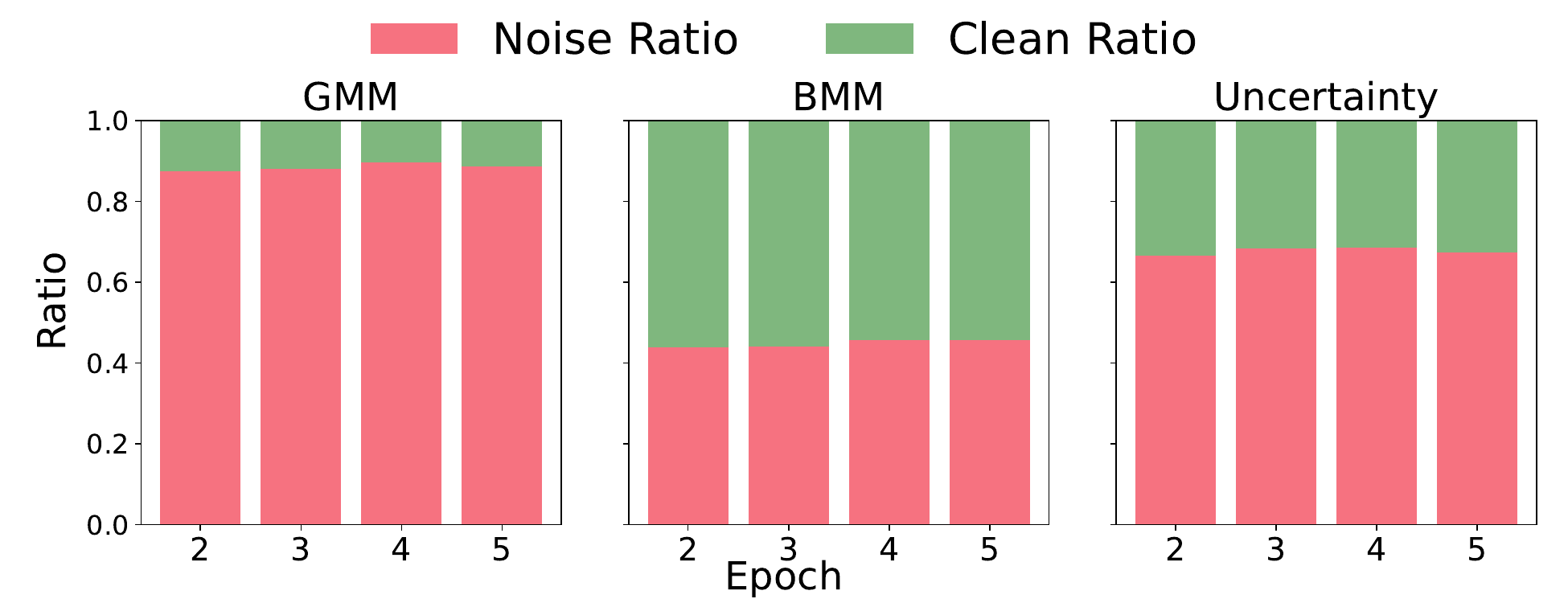} 
    \vspace{-0.1in}
    \caption{Proportions of true clean and true noisy samples within the set of samples classified as noisy by different divide strategies on GTA-UAV (cross-area, 30\% noise).}
    \vspace{-0.1in}
    \label{fig:Divide Method}
\end{figure}

Most NC frameworks utilize three partitioning approaches~\cite{ncr, Bicro, UGNCL}. Unlike conventional NC methods, PAUL targets more precise identification of noisy samples for effective augmentation and EDL. We apply different partition schemes on GTA-UAV (cross-area, 30\% noise) and record the ratio of true clean and noisy samples within the selected noisy sample set in each iteration. As seen in \cref{fig:Divide Method}, the GMM-based strategy achieves superior results, justifying our adoption of GMM for co-divide.

\subsubsection{Impact of Guidance Loss Function in Mask Generation}
\label{sec: selection}
Gradient-based mask generation for targeted augmentation is widely used~\cite{xu2022masked}, commonly guided by task loss. However, $\ell_{InfoNCE}$ guided gradients may bias toward already well-learned samples, limiting gains where improvement is most needed. Motivated by this, we compare $\mathcal{L}_\mathrm{EDL}$ guided, uncertainty guided, and $\ell_{InfoNCE}$ guided strategies for key region identification.

\begin{figure}[!htbp]
    \centering    
    \includegraphics[width=0.69\linewidth]{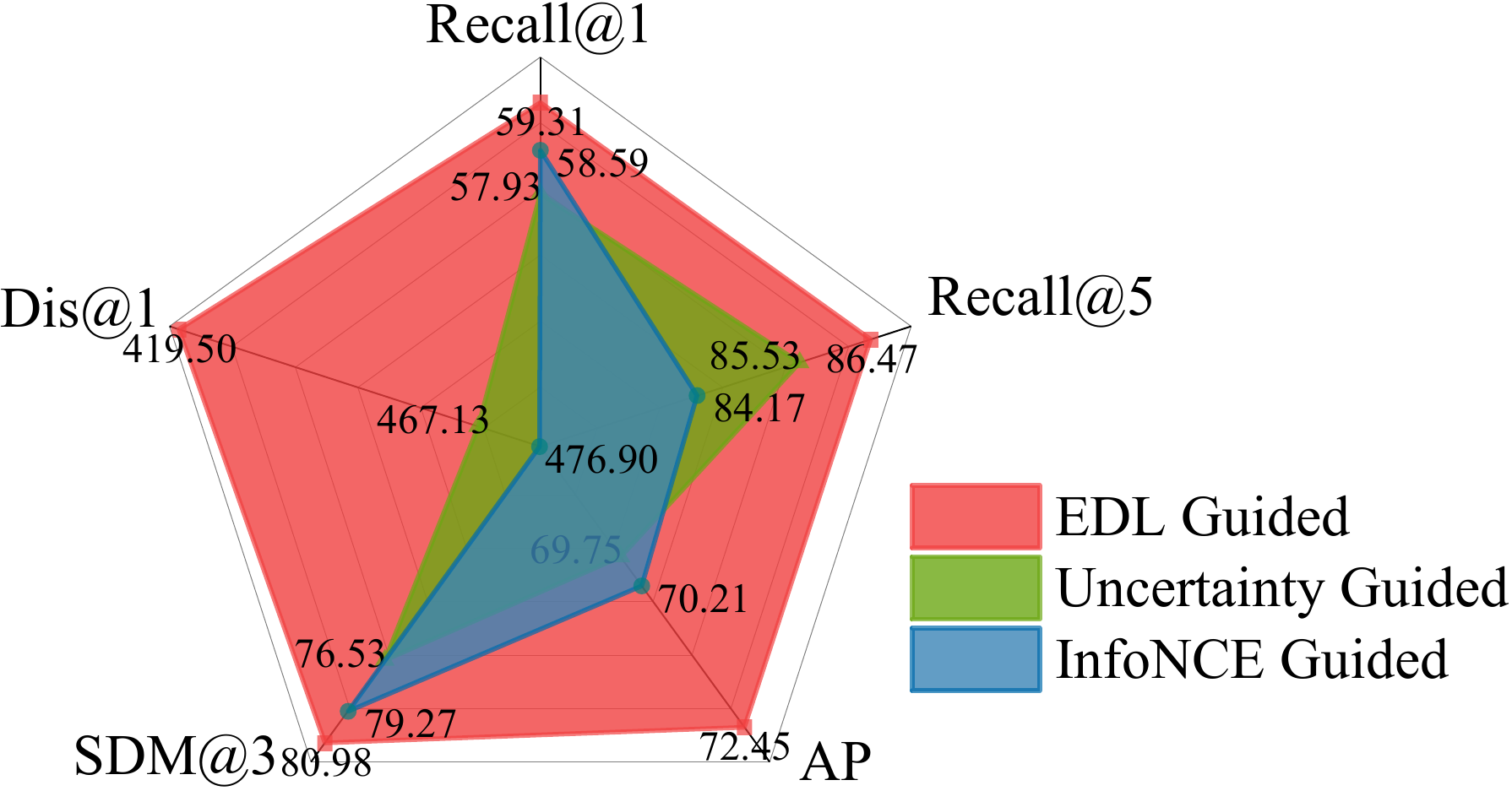} 
    \caption{Performance comparison of different augmentation strategies for mask generation on GTA-UAV (cross-area, 30\% noise).}
    \label{fig:Loss Function}
    \vspace{-0.15in}
\end{figure}

Empirical results (\cref{fig:Loss Function}) on GTA-UAV (cross-area, 30\% noise) demonstrate that $\mathcal{L}_\mathrm{EDL}$ guidance achieves superior performance. We attribute this to EDL's robust uncertainty quantification, which is less affected by sample difficulty distribution within a batch and thus better guides effective augmentation in noisy settings.

\subsubsection{Effectiveness of Augmentation Strategy}
\begin{figure}[htbp]
    \centering    
    \includegraphics[width=0.9\linewidth]{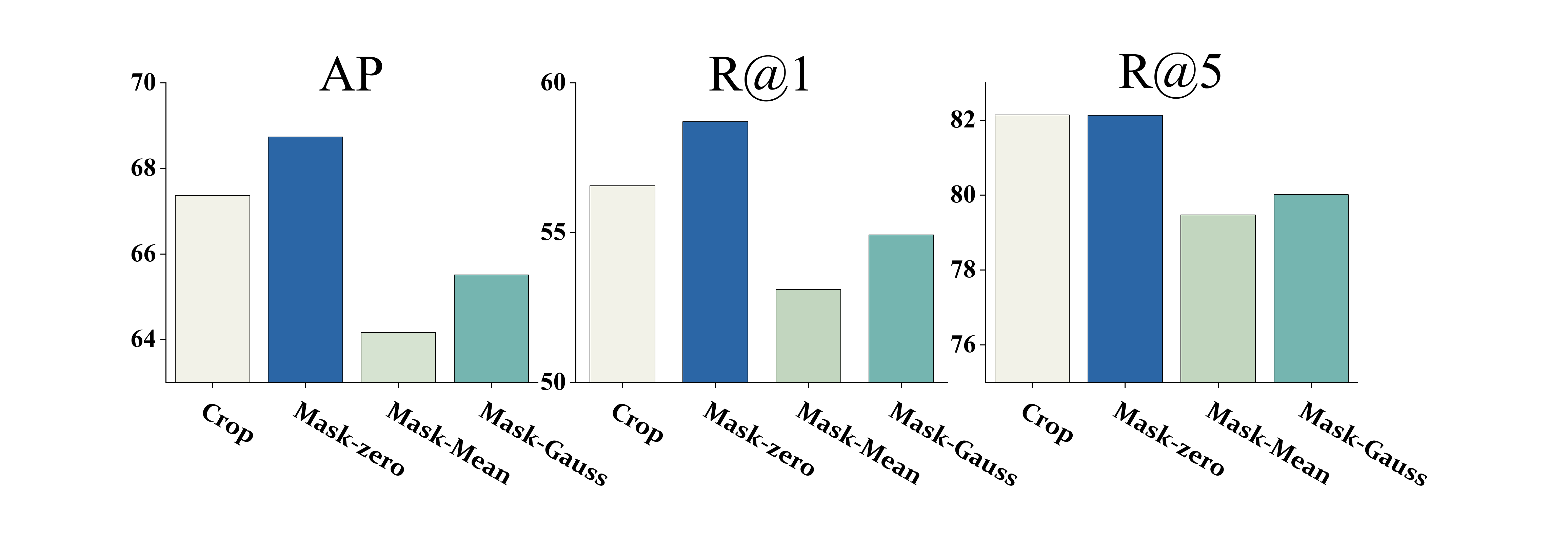} 
    \caption{Performance comparison of different augmentation strategies for mask generation on the GTA-UAV dataset (cross-area, 30\% noise).}
    \label{fig:Augmentation Method}
    \vspace{-0.1in}
\end{figure}

We systematically compare four augmentation strategies: local magnification, zero masking, Gaussian noise masking, and mean-value masking~\cite{liu2025og, lao2025boosting}. As shown in \cref{fig:Augmentation Method}, direct masking achieves the best results. Local magnification may introduce geometric distortions, while noise or mean-value masking can introduce additional errors, potentially impeding learning. In contrast, direct masking provides effective, artifact-free focus on challenging regions, which is consistent with our augmentation stratege design.

\subsubsection{Case Study}

\begin{figure}[htbp]
    \centering
    \includegraphics[width=0.99\linewidth]{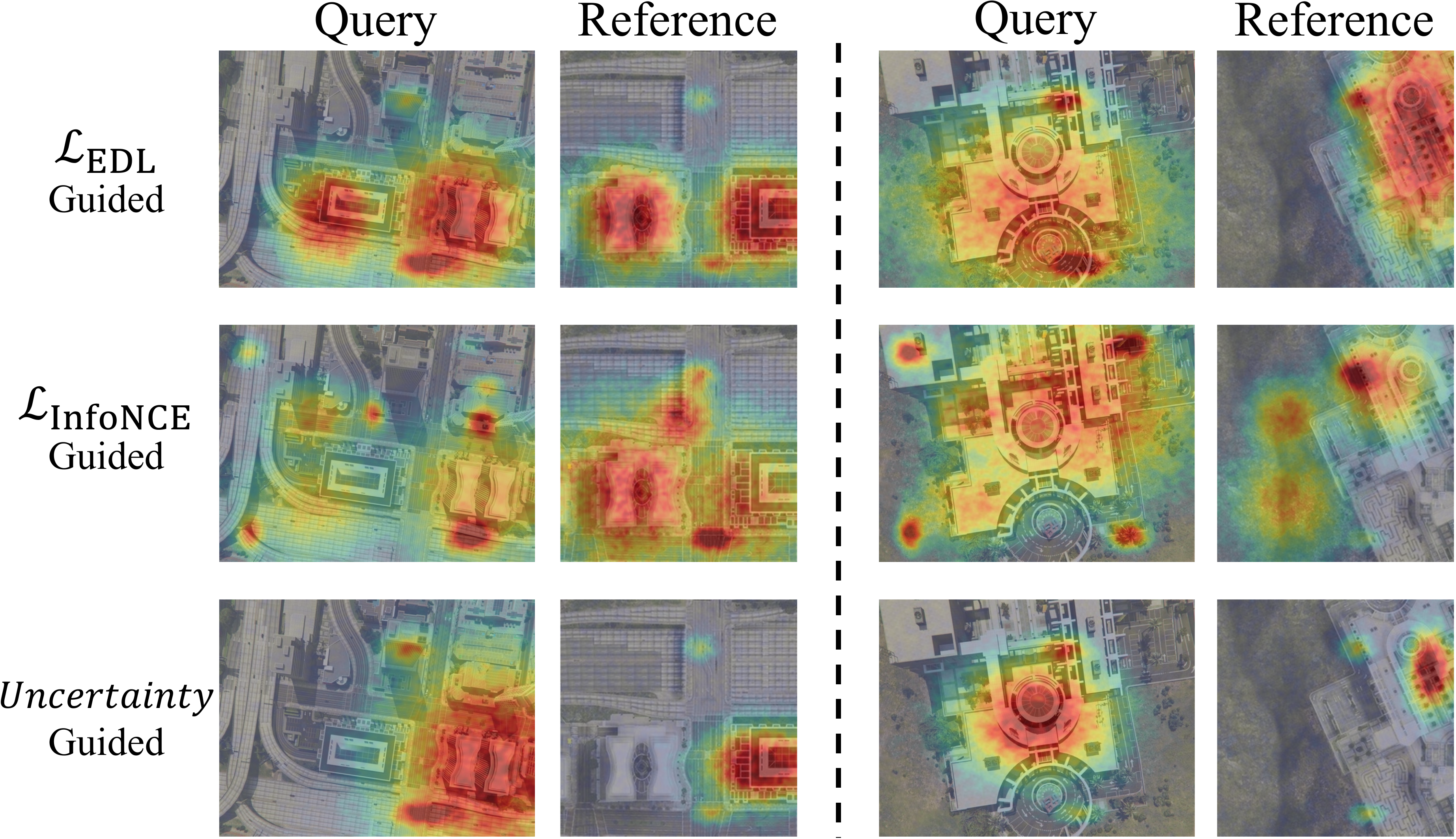} 
    \caption{Grad-CAM attention map visualizations for representative samples under different guidance loss function. }
    \label{fig:Attention Map}
    \vspace{-0.15in}
\end{figure}

To further substantiate the effectiveness of $\mathcal{L}_\mathrm{EDL}$ based guidance mask generation, we present Grad-CAM attention maps for representative samples. As shown in \cref{fig:Attention Map}, the attention maps generated using $\ell_{InfoNCE}$ guided gradients exhibit considerable deviation from the true critical regions, while those guided by uncertainty provide only partial coverage. In contrast, $\mathcal{L}_\mathrm{EDL}$ produces attention maps that comprehensively highlight regions of interest, verifying the precise alignment of consistent features across UAV-Satellite views. These observations provide additional empirical evidence for the superiority of PAUL.

\section{Conclusion}

This paper investigates the Noisy Correspondence in Cross-View Geo-Localization (NC-CVGL), a common challenge in CVGL stemming from annotation errors and spatial misalignment. Since existing methods lack robustness against the partial correspondence in noisy samples, we propose the PAUL framework. PAUL integrates evidential deep learning to model sample-level uncertainty, effectively managing noisy samples rather than discarding them. Our approach significantly enhances reliability across multiple datasets, successfully bridging the gap between theoretical benchmarks and practical applications. Future research will focus on constructing large-scale datasets with real-world noise and evaluating these methods on actual UAV platforms

\section*{Acknowledgments}
\vspace{-0.1in}
This work was supported by the National Science Foundation for Young Scientists of China(Grant No. 62502530), Hunan Province Key Research and Development Program (Grant No. 2025QK3004), the NUDT Foundation (Grant No. 25-ZZCX-JDZ-39, ZK24-27), and the Foundation of National Key Laboratory of Information Systems Engineering (Grant No. WDZC20265290415).
\vspace{-0.1in}

{
    \small
    \bibliographystyle{ieeenat_fullname}
    \bibliography{main}
}


\end{document}